%% file: main.tex
\documentclass[pdflatex,sn-mathphys-num,iicol]{sn-jnl}

\usepackage{graphicx}
\usepackage{multirow}
\usepackage{amsmath,amssymb,amsfonts}
\usepackage{amsthm}
\usepackage[title]{appendix}
\usepackage{xcolor}
\usepackage{textcomp}
\usepackage{booktabs}
\usepackage{algorithm}
\usepackage{algorithmicx}
\usepackage{algpseudocode}

\graphicspath{{figures/}}
\newcommand{\eps}{\varepsilon}

\DeclareMathOperator*{\argmax}{arg\,max}

\begin{document}

\title[Anchored Rank-Adversarial Fine-Tuning for VPR]{Anchored Rank-Adversarial Fine-Tuning of Pretrained Visual Place Recognition Models}

\author*[1]{\fnm{Mostafa} \sur{Kira}}\email{m.kira@innopolis.university}
\author[1]{\fnm{Bader} \sur{Rasheed}}\email{b.rasheed@innopolis.university}
\author[1]{\fnm{Manuel} \sur{Mazzara}}\email{m.mazzara@innopolis.ru}

\affil*[1]{\orgdiv{Faculty of Computer Science and Engineering}, \orgname{Innopolis University}, \orgaddress{\city{Innopolis}, \country{Russia}}}

\abstract{\input{sections/00_abstract}}


\maketitle

\input{sections/01_introduction}
\input{sections/02_related_work}
\input{sections/03_method}
\input{sections/04_setup}
\input{sections/05_results}
\input{sections/06_discussion}
\input{sections/07_conclusion}

\backmatter
\input{sections/08_declarations}

\begin{appendices}
\input{sections/09_appendix}
\end{appendices}

\bibliography{references}

\end{document}

%% file: sections/00_abstract.tex
\unboldmath% sn-jnl leaves \boldmath on after the abstract heading
Visual place recognition (VPR) localises a camera by retrieving similar geo-tagged images,
and current foundation-model VPR models are accurate across cities, seasons and
illumination, but not robust to adversarial perturbations: under a white-box attack on the ranking at
an imperceptible budget of $1/255$, SuperVLAD and BoQ keep at most 0.2\% top-1 recall
(R@1) on four benchmarks. We propose anchored
rank-adversarial fine-tuning (ARAFT), which fine-tunes a released model with a listwise
rank loss on each adversarial query's hardest positive and an anchor to the clean
descriptor of the frozen pretrained model. Checkpoints are selected under a clean-accuracy
budget fixed in advance, and the full trade-off is reported. Within five validation R@1
points of clean cost, plain adversarial training has no admissible checkpoint, whereas
ARAFT raises robust R@1 from zero to 47.3 on SuperVLAD and 55.3 on BoQ; plain adversarial
training exceeds this only at about three times the clean cost. Against unbudgeted plain adversarial training on the test sets, ARAFT matches its robust R@1 at $2/255$ with about ten more clean R@1 points. At $4/255$, plain adversarial training is 6--9 robust R@1 points ahead, while ARAFT still keeps its ten-point clean advantage. Developed on SuperVLAD, ARAFT transfers to BoQ
without re-tuning, and its robustness holds against stronger optimisers, with no sign of
gradient masking.

%% file: sections/01_introduction.tex
\section{Introduction}\label{sec:introduction}

Visual place recognition (VPR) estimates where an image was taken by retrieving the most
similar images from a database of geo-tagged references. It provides the coarse
localisation and loop-closure candidates on which mobile robots, autonomous vehicles and
augmented-reality systems build their metric pose estimates. Modern VPR models encode each
image into a global descriptor with a pretrained backbone and a learned aggregation layer,
and rank the database by descriptor distance~\cite{arandjelovic2016netvlad,berton2022benchmark}.
The most accurate current models build on the self-supervised DINOv2
backbone~\cite{oquab2024dinov2} with aggregators such as SALAD~\cite{izquierdo2024salad},
BoQ~\cite{alibey2024boq} and SuperVLAD~\cite{lu2024supervlad}, are trained on large
street-view collections~\cite{alibey2022gsvcities}, and are deployed as released
checkpoints.

Deep retrieval models are vulnerable to adversarial perturbations: small, crafted changes
to a query image can push its true matches down the ranking or pull a chosen wrong image
to the top~\cite{tolias2019tma,zhou2020advrank}. Malone et al.~\cite{malone2025vprattacks}
showed that VPR models, including a DINOv2-based one, are vulnerable to attacks that alter
the query features, and proposed
to detect them; no defense has been trained for VPR. We find that the gap is severe for
current models. Under a white-box attack on the ranking, both models we study keep at
most 0.2\% top-1 recall on four benchmarks at a budget of $1/255$.
For a system that relies on VPR to localise, this means an imperceptible perturbation of
the camera image is enough to make it report the wrong place.

Existing defenses do not address this setting. Adversarial training for retrieval and deep
metric learning~\cite{zhou2020advrank,zhou2022robustdml,zhou2024tpami,tian2024catride}
trains ResNet-18 or smaller networks on retrieval benchmarks and loses about 15--26 points
of clean R@1 on CUB and SOP. Adversarial fine-tuning of foundation
encoders~\cite{mao2023tecoa,schlarmann2024fare,wang2024pmgaft,khodabandeh2025lore}
preserves clean accuracy better, but it targets classification and zero-shot transfer, its
objectives do not reason about rankings, and it has not been evaluated on place
recognition.

We propose \emph{anchored rank-adversarial fine-tuning} (ARAFT), which fine-tunes a
pretrained VPR model end to end on three terms: its original metric-learning loss on clean
images; a listwise rank loss that requires an adversarially perturbed query to rank its
\emph{hardest} positive above the nearest negatives; and an anchor that keeps the
perturbed query's descriptor close to the clean descriptor of the frozen pretrained model.
Because adversarial fine-tuning trades clean accuracy for robustness and the trade-off
moves during training, we select checkpoints under an explicit clean-accuracy budget, fixed
before the experiments, and report the full trade-off curve rather than a single operating
point.

We develop ARAFT on SuperVLAD and apply it unchanged to BoQ, and compare it with plain
adversarial training that uses the same attack, positives and schedule but no anchor.
Within a 5-point clean budget on the validation set, plain adversarial training has no
usable checkpoint on either model, whereas ARAFT reaches a robust R@1 of 47.3 (SuperVLAD)
and 55.3 (BoQ). Plain adversarial training exceeds this robustness only after giving up
about three times as much clean accuracy. On the test sets, ARAFT matches plain
adversarial training at its most robust checkpoint at $2/255$ while keeping about ten more
clean R@1 points, and it is less robust at $4/255$ and beyond. Its robustness holds against
stronger optimisers, with no sign of gradient masking. The clean cost under strong distribution shift, up to 24 points
on SPED and Nordland, remains the main limitation.

Our contributions are:
\begin{itemize}
  \item a threat model and evaluation protocol for adversarial robustness of VPR, with
    rank attacks, robustness curves and gradient-masking checks, and the finding that current foundation-model
    VPR models have no robustness under it;
  \item ARAFT, a rank-aware, anchored adversarial fine-tuning method for pretrained VPR
    models, with checkpoint selection under an explicit clean-accuracy budget;
  \item an evaluation on two architectures and four benchmarks, with paired significance
    tests and ablations, showing that ARAFT is the only one of the compared methods that
    provides robustness within a small clean budget, that it transfers to a second
    architecture without re-tuning, and that protecting the hardest positive and anchoring
    to the pretrained model are what make it work.
\end{itemize}

%% file: sections/02_related_work.tex
\section{Related work}\label{sec:related_work}

\paragraph{Visual place recognition.} VPR is commonly cast as image retrieval with global
descriptors. NetVLAD~\cite{arandjelovic2016netvlad} made the VLAD aggregation of local CNN
features trainable end to end, and later work improved training data and
objectives~\cite{alibey2022gsvcities,berton2023eigenplaces} and aggregation, for example
with the feature-mixing MixVPR~\cite{alibey2023mixvpr}. Recent state-of-the-art models
replace the CNN backbone with DINOv2~\cite{oquab2024dinov2}: SALAD~\cite{izquierdo2024salad}
aggregates its patch features by optimal transport, BoQ~\cite{alibey2024boq} with a bag of
learnable queries, and SuperVLAD~\cite{lu2024supervlad} with a compact VLAD layer and a
cross-image encoder. These models are typically trained on GSV-Cities with a
multi-similarity loss~\cite{wang2019multisimilarity} and evaluated on benchmarks that span
urban, seasonal and day--night change~\cite{berton2022benchmark}. Their robustness is
studied with respect to such natural changes, not to adversarial perturbations.

\paragraph{Adversarial attacks on retrieval and VPR.} Adversarial examples were first
studied for classification~\cite{szegedy2014intriguing,goodfellow2015fgsm,madry2018robustness}, where
AutoAttack~\cite{croce2020autoattack} has become the standard for reliable evaluation.
For retrieval, Tolias et al.~\cite{tolias2019tma} showed that a query can be perturbed to
retrieve the results of an unrelated image, and Zhou et al.~\cite{zhou2020advrank,zhou2024tpami}
formalised attacks that raise or lower the rank of chosen candidates by perturbing the
query or the database. For VPR, Malone et al.~\cite{malone2025vprattacks} showed that
NetVLAD-, AP-GeM- and DINOv2-based models are vulnerable to feature-space attacks that
copy reference or query content and, to a lesser degree, to FGSM attacks transferred from
another model, and proposed detecting attacks as a safeguard for robot
navigation. They train no defense. We study the white-box, query-side threat model in
which the attacker optimises against the ranking itself, and train models to resist it.

\paragraph{Adversarially robust retrieval and metric learning.} Defenses for ranking
models adapt adversarial training~\cite{madry2018robustness} to triplet or ranking
losses. Zhou et al.~\cite{zhou2020advrank,zhou2024tpami} proposed embedding-shifting and
anti-collapse objectives; hardness manipulation~\cite{zhou2022robustdml} perturbs
training triplets to a chosen hardness, which makes adversarial training efficient; and
CA-TRIDE~\cite{tian2024catride} decouples the perturbation of anchors and candidates and
adapts it to the model's state of collapse. These methods train ResNet-18 or smaller
networks on fine-grained retrieval benchmarks (MNIST, Fashion-MNIST, CUB, Cars, SOP), the
later ones from ImageNet initialisation at budgets of $8/255$, and lose about 15--26 points
of clean R@1 on CUB and SOP. They do not start from a strong pretrained retrieval
model, do not consider geographic retrieval under distribution shift, and do not bound
the clean cost. Our plain adversarial-training baseline follows this line with a hinge
ranking loss, and our collapse guard is motivated by the collapse these works report.

\paragraph{Robust fine-tuning of pretrained encoders.} A second line adversarially
fine-tunes large pretrained models while trying to keep their clean ability.
TeCoA~\cite{mao2023tecoa} fine-tunes CLIP with a supervised contrastive adversarial loss;
PMG-AFT~\cite{wang2024pmgaft} adds a term that keeps the outputs close to the original
model's; FARE~\cite{schlarmann2024fare} fine-tunes the image encoder without labels so that
adversarial embeddings stay close to the original clean embeddings; and
LORE~\cite{khodabandeh2025lore} turns this proximity into an explicit constraint and also
applies it to DINOv2. TWINS~\cite{liu2023twins} keeps frozen statistics of the pretrained
network during adversarial fine-tuning, and Ngnaw{\'e} et al.~\cite{ngnawe2026epsilon}
show that fine-tuning a non-robust model directly at the target budget transfers poorly
and that ramping the budget avoids this. Croce et al.~\cite{croce2025rclipf} evaluate
such robust encoders on image retrieval. These methods target classification, zero-shot
transfer or general-purpose features; their losses do not involve rankings, and none is
evaluated on place recognition. ARAFT borrows FARE's anchor and the budget ramp from this
line and combines them with a rank-aware loss on the hardest positive, applied to a
complete VPR model.

%% file: sections/03_method.tex
\section{Method}\label{sec:method}

We consider a pretrained VPR model $f_0$ that maps an image to an $\ell_2$-normalised
global descriptor, and fine-tune it into a model $f_\theta$, initialised at $f_0$, whose
top-1 retrieval survives perturbations of the query image while its clean retrieval stays
close to that of $f_0$. This section defines the threat model (Sect.~\ref{sec:threat}),
the rank attacks used for both training and evaluation (Sect.~\ref{sec:attacks}), the
proposed objective (Sect.~\ref{sec:araft}), the plain adversarial-training baseline
(Sect.~\ref{sec:plain_at}) and how checkpoints are selected (Sect.~\ref{sec:selection}).

\subsection{Problem setting and threat model}\label{sec:threat}

\paragraph{Retrieval.} A database of geo-tagged images $\{d_j\}_{j=1}^{M}$ is encoded
once, offline, into descriptors $f(d_j)$. A query image $x$ is localised by ranking the
database by the Euclidean distance $\|f(x)-f(d_j)\|_2$. The positives $\mathcal{P}(x)$
are the database images taken within a fixed distance of the query's true position, and
the query is retrieved correctly at rank one (R@1) if its nearest database descriptor
belongs to $\mathcal{P}(x)$.

\paragraph{Adversary.} The adversary has white-box access to the model weights and to
the clean database descriptors, and perturbs only the query. The database is built
offline and is read-only, as in prior work on VPR attacks~\cite{malone2025vprattacks};
attacks on database images are out of scope. The perturbation $\delta$ lies in an
$\ell_\infty$ ball, $\|\delta\|_\infty\le\eps$, applied in the ImageNet-normalised input
space in which the model operates. A normalised budget $\eps_n$ corresponds to a pixel
budget of $\eps_n\sigma_c$ in channel $c$, with $\sigma=(0.229,0.224,0.225)$; we quote
the largest of these, $\eps=\eps_n\cdot 0.229$, in units of $1/255$, so a stated budget
of $4/255$ bounds every channel by $4/255$. Physical and patch attacks are out of scope,
and all robustness we report is empirical rather than certified.

\paragraph{Attack goal.} An attack succeeds when no positive remains at rank one.

\subsection{Rank attacks}\label{sec:attacks}

Attacks on a retrieval model have no class logits to maximise; we attack the ranking
directly. For a query $x$ let $\mathcal{P}$ be its positive descriptors and $\mathcal{N}$
the descriptors of its $k=5$ nearest non-positive database images under the clean query.
The \emph{rank score} of a perturbed query $x'=x+\delta$ is
\begin{equation}
  s(x') = m + \min_{p\in\mathcal{P}}\|f(x')-p\|_2 - \min_{n\in\mathcal{N}}\|f(x')-n\|_2,
  \label{eq:rank_score}
\end{equation}
with margin $m=0.1$. The query is pushed out of rank one once $s(x')>m$. Following the
ranking attacks of Zhou et al.~\cite{zhou2020advrank}, \emph{rank-PGD} ascends $s$ by $T$
signed-gradient steps of size $\alpha=2\eps/T$, each followed by projection onto the
$\eps$-ball and clipping to the valid image range, and returns the iterate with the
highest score. It ascends the raw score rather than its hinge $\max(0,s)$: the hinge has
zero gradient for a confidently correct query, which would leave exactly the queries
that matter most unattacked. The first run starts from the clean image and further
restarts from a uniformly random point in the ball. \emph{Rank-APGD} uses the same
objective with a per-query step size that is halved whenever the best score has not
improved over the last $T/4$ iterations, a simplified form of the step-size control of
APGD~\cite{croce2020autoattack} without its momentum, restarts from the best point or
non-uniform checkpoints.

In every case success is measured by the true ranking of the perturbed query against the
full database, not by the surrogate score.

\subsection{Anchored rank-adversarial fine-tuning}\label{sec:araft}

ARAFT (\emph{Anchored Rank-Adversarial Fine-Tuning}) fine-tunes the whole network on
three terms: the model's original metric-learning loss on clean images, a listwise rank
loss on adversarial queries computed against the hardest positive, and an anchor that
ties adversarial descriptors to the frozen pretrained model. Algorithm~\ref{alg:araft}
summarises one training step.

\paragraph{Batch and targets.} A training batch holds $B$ places with $K$ images each.
For every place, the first image is the query $x$. Its clean descriptor, and those of all
other batch images, are computed with the current model without gradients.
For SuperVLAD the query is encoded on its own and the other batch images in one
joint pass (Sect.~\ref{sec:models}). The
\emph{hardest positive} $p^+$ is the other image of the same place whose descriptor is
farthest from the query's, and $\mathcal{N}$ holds the five images of other places
nearest to the query. The training attack is rank-PGD (Sect.~\ref{sec:attacks}) with
$\mathcal{P}=\{p^+\}$, budget $\eps_{\mathrm{train}}$ and $T=5$ steps, run against the
current model; it yields the adversarial query $x'$.

\paragraph{Objective.} With $\mathcal{B}$ the clean batch, the loss is
\begin{equation}
  \begin{aligned}
  \mathcal{L} = \mathcal{L}_{\mathrm{clean}}(\mathcal{B})
  &+ \lambda_{\mathrm{rank}}\,\mathcal{L}_{\mathrm{rank}}\big(f_\theta(x');\,p^+,\mathcal{N}\big)\\
  &+ \lambda_{\mathrm{anc}}\,\big\|f_\theta(x')-f_0(x)\big\|_2^2 ,
  \end{aligned}
  \label{eq:araft}
\end{equation}
averaged over the queries of the batch. $\mathcal{L}_{\mathrm{clean}}$ is the
multi-similarity loss with its pair miner~\cite{wang2019multisimilarity}, the loss with
which the pretrained models were trained, applied unchanged to the clean batch.

\paragraph{Listwise rank loss.} Writing $q=f_\theta(x')$ and $d(\cdot,\cdot)$ for the
Euclidean distance, the soft rank of the hardest positive among $\{p^+\}\cup\mathcal{N}$ is
\begin{equation}
  \tilde r(q) = 1 + \sum_{n\in\mathcal{N}} \sigma\!\left(\frac{d(q,p^+)-d(q,n)}{\tau}\right),
\end{equation}
which tends to its integer rank as $\tau\to0$. The loss penalises a soft rank above $K$,
\begin{equation}
  \mathcal{L}_{\mathrm{rank}} = 4\tau^2\,\mathrm{softplus}\!\left(\frac{\tilde r(q)-K-\tfrac12}{\tau}\right),
  \label{eq:listwise}
\end{equation}
with $K=1$ and $\tau=0.05$. It belongs to the Smooth-AP family~\cite{brown2020smoothap}
but targets top-1 recall, the metric VPR is judged on, instead of average precision. The
offset of one half separates an acceptable rank $K$ from a violating rank $K+1$.
The factor $4\tau^2$ cancels the two factors $1/\tau$ and the maximum sigmoid
slope of $1/4$, so the derivative of $\mathcal{L}_{\mathrm{rank}}$ with respect to each
negative distance $d(q,n)$ is at most one in magnitude, the slope of the hinge in
Eq.~\eqref{eq:plain_at}. The positive distance $d(q,p^+)$ enters all $|\mathcal{N}|$
terms of $\tilde r(q)$, so the derivative with respect to it can reach $|\mathcal{N}|=5$.
The factor therefore sets the scale of the loss but does not bound the rank term
relative to $\mathcal{L}_{\mathrm{clean}}$; their balance is set by
$\lambda_{\mathrm{rank}}$.

\paragraph{Why the hardest positive.} A query is retrieved correctly when \emph{any}
positive is ranked first. Defending only the closest positive is therefore sufficient
for correctness, but the resulting loss is close to zero as soon as one positive is safe,
so most queries contribute no gradient. Requiring the hardest positive to outrank the
negatives is the tightest version of the same constraint, and it keeps every attacked
query in the loss. Section~\ref{sec:results}
compares the two choices.

\paragraph{Anchor.} The last term of Eq.~\eqref{eq:araft} is the unsupervised
embedding-alignment loss of FARE~\cite{schlarmann2024fare}: the descriptor of the
adversarial query under the fine-tuned model must stay close to the descriptor of the
\emph{clean} query under the frozen pretrained model $f_0$. Because descriptors have unit
norm, the term lies in $[0,4]$. It serves two purposes at once. For robustness, it asks the
attacked query to land where the pretrained model placed the clean one, independently of
the mined positives and negatives. For clean accuracy, it ties $f_\theta$ to the released
model; at zero perturbation it reduces to $\|f_\theta(x)-f_0(x)\|_2^2$, a direct penalty on
clean drift. It is applied to the attacked queries only.

\paragraph{Training schedule and collapse guard.} Robust fine-tuning of a non-robust
pretrained model loses clean accuracy when trained at the full budget from the
start~\cite{ngnawe2026epsilon}. ARAFT therefore trains the first epoch without attacks and
then ramps the training budget linearly from $0.1\,\eps_{\mathrm{train}}$ to
$\eps_{\mathrm{train}}$ over five epochs. Validations during the ramp still update the
saved checkpoints but do not count toward learning-rate or early-stopping patience. To
detect representation collapse, a known failure of adversarial metric
learning~\cite{zhou2022robustdml,tian2024catride}, we measure after each validation the
mean overlap between the 10 nearest neighbours of 256 fixed validation images under
$f_\theta$ and under $f_0$, and abort training if it falls below $0.15$.

\begin{algorithm}[t]
\caption{One ARAFT training step}\label{alg:araft}
\begin{algorithmic}[1]
\Require batch of $B$ places $\times$ $K$ images; model $f_\theta$;
\Statex \hspace{\algorithmicindent}frozen $f_0$; budget $\eps$ of this epoch
\State compute clean descriptors of all batch images with $f_\theta$, without gradients
\For{each place $i$}
  \State $x_i \gets$ first image; $p_i^+ \gets$ farthest same-place descriptor
  \State $\mathcal{N}_i \gets$ 5 nearest other-place descriptors
\EndFor
\State $x'_i \gets$ rank-PGD$(f_\theta, x_i, \{p_i^+\}, \mathcal{N}_i, \eps, T{=}5)$
\State $\ell_i \gets \lambda_{\mathrm{rank}}\mathcal{L}_{\mathrm{rank}}(f_\theta(x'_i); p_i^+, \mathcal{N}_i)$
\Statex \hspace{\algorithmicindent}$\phantom{\ell_i \gets}{} + \lambda_{\mathrm{anc}}\|f_\theta(x'_i)-f_0(x_i)\|_2^2$
\State $\mathcal{L} \gets \mathcal{L}_{\mathrm{clean}}(\text{batch}) + \frac{1}{B}\sum_i \ell_i$
\State update $\theta$ with a gradient step on $\mathcal{L}$
\end{algorithmic}
\end{algorithm}

\subsection{Plain adversarial training baseline}\label{sec:plain_at}

The baseline is adversarial training with a ranking loss in the style of Madry et
al.~\cite{madry2018robustness} and of robust deep metric
learning~\cite{zhou2020advrank,zhou2022robustdml}:
\begin{equation}
  \begin{aligned}
  \mathcal{L}_{\mathrm{AT}} = \mathcal{L}_{\mathrm{clean}}(\mathcal{B})
  &+ \lambda_{\mathrm{rank}}\max\!\big(0,\\
  &\quad m + d(q,p^+) - \min_{n\in\mathcal{N}} d(q,n)\big).
  \end{aligned}
  \label{eq:plain_at}
\end{equation}
It uses the same hardest positive, negatives, training attack, schedule, collapse guard
and checkpoint selection as ARAFT, and differs from it in having no anchor and a hinge in
place of the listwise surrogate. Section~\ref{sec:results} reports the effect of each
difference separately.

\subsection{Checkpoint selection under a clean budget}\label{sec:selection}

Adversarial fine-tuning trades clean accuracy for robustness, and the trade-off moves
during training. Published robust-retrieval defenses report clean costs of about 15--26 R@1
points on CUB and SOP~\cite{zhou2020advrank,zhou2022robustdml,tian2024catride}, and none
selects checkpoints under an explicit clean-accuracy budget.
A deployed VPR system has a limit on how much clean accuracy it can give up, so we select
checkpoints under an explicit clean budget and also report the full trade-off.

Every three epochs, the model is validated on a fixed set of validation queries: clean
R@1 $C_e$, and R@1 under rank-PGD with 10 steps at $2/255$ and at $4/255$. The robust
score $S_e$ is the mean of the two attacked R@1 values. With $C_0$ the clean R@1 of the
pretrained model, each run keeps
\begin{itemize}
  \item the \emph{most robust} checkpoint, $e^\star=\argmax_e S_e$; and
  \item for each clean budget $b$, the \emph{budget-$b$} checkpoint
    $e_b=\argmax_{e:\,C_0-C_e\le b} S_e$, which is the pretrained model itself if no
    trained epoch stays within the budget.
\end{itemize}
The headline budget, $b=5$ R@1 points, was fixed before any experiment. It is well above
the variation of clean fine-tuning alone and below the clean cost of published ranking
defenses. The budget is a rule applied on the validation set; the clean cost realised on
the test sets can differ and is always reported next to the robust numbers. Test sets are
used only for final models.

%% file: sections/04_setup.tex
\section{Experimental setup}\label{sec:setup}

\subsection{Models}\label{sec:models}

We developed and tuned the method on one VPR model and then applied the same recipe,
without re-tuning, to a second model with a different aggregator.

\paragraph{SuperVLAD.} SuperVLAD~\cite{lu2024supervlad} aggregates DINOv2
ViT-B/14~\cite{oquab2024dinov2} patch features with a VLAD-style layer of four clusters
and one ghost cluster, followed by a cross-image encoder. We start from the released
checkpoint trained on GSV-Cities and use $322\times322$ inputs for training and testing.
The cross-image encoder attends across all images of a batch, so a descriptor depends on
the other images it is encoded with. An attack evaluated in batches would then succeed or
fail per batch rather than per query. In all attacks, validation and testing we therefore
encode every query and database image on its own, passing a sequence of length one to
the cross-image encoder. During training the clean loss keeps the original batch-level
encoding, and adversarial queries and anchor descriptors are encoded individually.
The positive and negative descriptors that serve as training targets are taken,
without gradients, from a batch-level pass over all 80 images of the batch; only the
query's clean descriptor, which is used to mine them, is encoded on its own. Training
targets are therefore encoded in a different context from the validation and test
databases, in which every image is encoded on its own.

\paragraph{BoQ.} BoQ~\cite{alibey2024boq} uses the same DINOv2 ViT-B/14 backbone with a
Bag-of-Queries aggregator and 12\,288-dimensional descriptors. We start from the released
GSV-Cities checkpoint and, like its public training code, train at $224\times224$ and test at
$322\times322$. BoQ encodes each image independently. All ARAFT and baseline settings are
those selected on SuperVLAD; only the starting checkpoint and the input resolution change.
BoQ therefore tests whether the recipe transfers to a new architecture without a new
hyperparameter search.

\paragraph{Compared models.} For each architecture we evaluate the released model, plain
adversarial training (Sect.~\ref{sec:plain_at}) and ARAFT, each at its budget-5 and most
robust checkpoints (Sect.~\ref{sec:selection}). On SuperVLAD we also train a second ARAFT
seed and a \emph{clean twin}, fine-tuned with the same data, depth and schedule but
without attacks. The twin separates the clean cost of fine-tuning itself from the cost of
robustness. Clean fine-tuning never improved on the pretrained model's validation R@1,
so the twin is evaluated at its last checkpoint.

\subsection{Training details}\label{sec:training}

All models are trained on GSV-Cities~\cite{alibey2022gsvcities} in batches of $B=20$
places with $K=4$ images each, 200 batches per epoch, with all network parameters
trainable. Table~\ref{tab:hparams} lists the settings, which are identical for ARAFT and
plain adversarial training and for both architectures. We use Adam with learning rate
$10^{-5}$, no weight decay, gradient clipping at norm 1, mixed precision and gradient
checkpointing. Training runs for at most 100 epochs. The learning rate is divided by 10
after three validations without improvement of the robust score, and training stops after
six. The headline training budget $\eps_{\mathrm{train}}=4/255$ matches the
budget used for robust fine-tuning of foundation encoders~\cite{schlarmann2024fare}.

\begin{table}[t]
\caption{Training settings, shared by ARAFT and plain adversarial training and by both
architectures except where noted.}\label{tab:hparams}
\footnotesize
\begin{tabular}{@{}p{0.4\columnwidth}p{0.56\columnwidth}@{}}
\toprule
Setting & Value \\
\midrule
Training data & GSV-Cities, $20$ places $\times$ $4$ images \\
Batches/epoch, max.\ epochs & $200$ / $100$ \\
Optimiser & Adam, lr $10^{-5}$, no weight decay \\
Trainable layers & all \\
Input size (train / test) & SuperVLAD $322$ / $322$; BoQ $224$ / $322$ \\
Training attack & rank-PGD $\ell_\infty$, $\eps=4/255$, $T=5$ \\
Budget ramp & 1 clean epoch, then $0.1\eps\to\eps$ over 5 epochs \\
Loss weights & $\lambda_{\mathrm{rank}}=1$, $\lambda_{\mathrm{anc}}=1$ (ARAFT) \\
Rank loss & listwise, $K=1$, $\tau=0.05$ (ARAFT); hinge (plain AT) \\
Margin $m$, negatives $|\mathcal{N}|$ & $0.1$ / $5$ \\
Validation & every 3 epochs, 2\,000 MSLS-val queries \\
LR decay / early stop & $\times0.1$ after 3 / stop after 6 validations \\
Collapse guard & abort if 10-NN overlap $<0.15$ \\
\botrule
\end{tabular}
\end{table}

\paragraph{Hyperparameter search.} All settings were chosen on SuperVLAD using MSLS-val
only, by a sequential one-factor-at-a-time search. Short screens of 12 epochs first fixed
the fine-tuning depth and learning rate (0, 4 or 8 frozen DINOv2 blocks; $10^{-5}$ or
$3\cdot10^{-6}$), the set of training attacks ($\ell_\infty$ rank-PGD alone, or together
with two perceptual attacks~\cite{laidlaw2021perceptual}) and the anchor weight
($\lambda_{\mathrm{anc}}\in\{0,0.1,1,10\}$). These screens used the closest positive as
the rank target. Full-length runs then compared the closest and the hardest positive, and
the anchor weight was re-checked with the hardest positive
($\lambda_{\mathrm{anc}}\in\{1,2\}$). The baseline received its own learning-rate search over the same two
values. Appendix~\ref{app:search} gives the full search.

\subsection{Datasets and metrics}\label{sec:data}

Models are validated on 2\,000 fixed queries of MSLS-val~\cite{warburg2020msls} and tested
on four benchmarks: MSLS, Pitts30k~\cite{torii2013pittsburgh,arandjelovic2016netvlad},
SPED~\cite{chen2017sped} and Nordland~\cite{sunderhauf2013nordland}
(Table~\ref{tab:datasets}). No test image is used for training or model selection. MSLS
shares its domain with the validation set, whereas Pitts30k, SPED and Nordland measure
behaviour under distribution shift, including seasonal and day--night change. We use the
dataset formats and the 25\,m positive threshold of the VPR benchmark of Berton et
al.~\cite{berton2022benchmark}.
The MSLS test set is the official test split (Athens, Bengaluru, Buenos Aires, Kampala,
Miami and Stockholm), disjoint from the validation cities (Copenhagen and San Francisco).
SPED and Nordland have no geographic coordinates; the benchmark format assigns synthetic
ones so that a SPED query has a single positive, the database image of the same place, and
a Nordland query has as positives the database frames within $\pm 10$ frames of it.

\begin{table}[t]
\caption{Evaluation datasets. Attacked queries are evenly spaced over all queries with
at least one positive. This holds for the test sets; the 2000 MSLS-val queries
are a fixed random sample of such queries, on which validation measures both clean and
attacked R@1.}\label{tab:datasets}
\begin{tabular}{@{}lrrr@{}}
\toprule
Dataset & Database & Queries & Attacked \\
\midrule
MSLS-val (val.) & 18\,871 & 11\,084 & 2\,000 \\
MSLS & 36\,558 & 25\,622 & 1\,000 \\
Pitts30k & 10\,000 & 6\,816 & 1\,000 \\
SPED & 607 & 607 & 607 \\
Nordland & 27\,592 & 27\,592 & 1\,000 \\
\botrule
\end{tabular}
\end{table}

\paragraph{Clean and attacked recall.} Clean R@1 is computed on all queries of each test
set. Running strong attacks on every query is too expensive (a single 20-step attack on
all of Nordland takes about 6.6 hours per model and budget), so attacked R@1 is computed
on at most 1\,000 queries per dataset, evenly spaced over the query list so that they
cover whole routes; SPED is attacked in full. All attacked numbers for a dataset use the
same queries for every model.

\paragraph{Summary metrics.} Besides R@1 at each budget we report the \emph{clean drop}:
the loss in clean R@1 relative to the pretrained model and, on SuperVLAD, relative to the
clean twin.
Per-query outcomes of two models on the same attacked queries are compared with the exact
McNemar test~\cite{mcnemar1947}.

\subsection{Attack protocol}\label{sec:protocol}

Table~\ref{tab:protocol} lists the evaluation attacks. The headline attack is rank-APGD
with 100 steps at $2/255$ and $4/255$; rank-PGD with 20 steps traces the robustness curve.
Both architectures are evaluated with these two attacks. The remaining attacks test, on
SuperVLAD, whether the robustness could be an artefact of gradient masking: attacked R@1
must not grow with $\eps$, must reach about zero at a very large budget, and must not
change when the optimiser gets five times more steps or three restarts.

\begin{table*}[t]
\caption{Evaluation attacks. All attacks are white-box and use the full model; budgets are
the largest per-channel pixel budget, with the normalised value $\eps_n$ in parentheses.}
\label{tab:protocol}
\footnotesize
\begin{tabular*}{\textwidth}{@{\extracolsep\fill}llllc@{}}
\toprule
Attack & Steps$\times$restarts & Budget & Purpose & Models \\
\midrule
Rank-APGD $\ell_\infty$ & $100\times1$ & $2/255$ ($0.03425$), $4/255$ ($0.0685$) & headline & both \\
Rank-PGD $\ell_\infty$ & $20\times1$ & $1,2,4,8/255$ (to $0.137$) & budget sweep & both \\
Rank-APGD $\ell_\infty$ & $500\times1$; $100\times3$ & $4/255$, MSLS & convergence & SuperVLAD \\
Rank-PGD $\ell_\infty$ & $20\times1$ & $\approx32/255$ ($0.548$) & large budget & SuperVLAD \\
\botrule
\end{tabular*}
\end{table*}

%% file: sections/05_results.tex
\section{Results}\label{sec:results}

\subsection{Pretrained VPR models are not robust}\label{sec:res_pretrained}

Both pretrained models lose all top-1 recall under the smallest budget we test. Rank-PGD
at $1/255$ reduces R@1 to 0.0 on all four test sets for SuperVLAD and to at most 0.2 for
BoQ, and every larger budget gives 0.0. A perturbation
below the visibility threshold is therefore enough to make a VPR system report the wrong
place.

\subsection{Clean--robust trade-off}\label{sec:res_tradeoff}

\begin{figure*}[t]
\centering
\includegraphics[width=\textwidth]{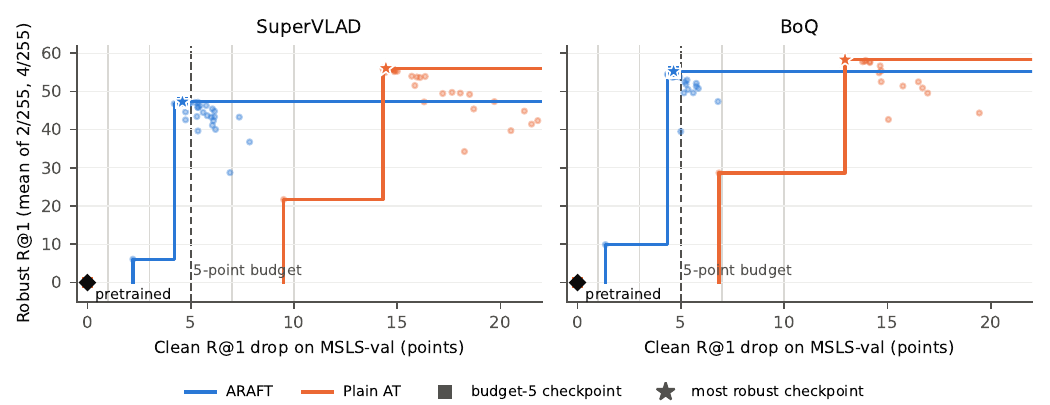}
\caption{Clean--robust trade-off on MSLS-val. Each dot is one validated epoch; the robust
score is the mean R@1 under rank-PGD at $2/255$ and $4/255$. Lines give the highest robust
score reached within each clean drop. Squares and stars mark the budget-5 and most robust
checkpoints; the dashed line is the 5-point budget, grey lines the other budgets.}
\label{fig:tradeoff}
\end{figure*}

Figure~\ref{fig:tradeoff} plots every validated epoch of every run by its clean cost and
its robust score on MSLS-val, together with the best robust score each method reaches
within a given clean cost. The two methods follow different paths from the first
adversarial epoch on. After three epochs, plain adversarial training has already lost 9.5
clean R@1 points on SuperVLAD and 6.9 on BoQ; ARAFT has lost 2.2 and 1.4. ARAFT's robust
score then rises while its clean drop stays between about 4 and 8 points, and its most
robust checkpoint lies inside the 5-point budget on both models: 47.3 robust R@1 at a
clean drop of 4.6 on SuperVLAD, and 55.3 at 4.7 on BoQ.

Plain adversarial training has no checkpoint inside the 5-point budget on either model, so
its budget-5 model is the unmodified pretrained one, with zero robustness. Even with a
10-point budget its best robust score is 21.7 on SuperVLAD and 28.7 on BoQ, less than half
of ARAFT's. Plain adversarial training first exceeds ARAFT's robust score at a clean drop
of 14.3 points on SuperVLAD and 13.0 on BoQ, and peaks at 56.1 and 58.3. Within the range
of clean costs that a deployed system can plausibly accept, ARAFT is therefore the only
one of the two methods that provides robustness; plain adversarial training buys more
robustness only at roughly three times the clean cost.

\subsection{Robustness on the test sets}\label{sec:res_main}

We compare the two methods on the test sets twice: under the 5-point clean budget, each
at its budget-5 checkpoint (Table~\ref{tab:budget}), and without the budget, each at its
most robust checkpoint (Table~\ref{tab:main}). On both models ARAFT's most robust
checkpoint is its budget-5 one, so its rows are the same in both tables.

\begin{table*}[t]
\caption{Test results under the 5-point clean budget, each method at its budget-5
checkpoint. Plain AT has no checkpoint within the budget, so its budget-5 model is the
unchanged pretrained model. Clean R@1 on all queries; R@1 under rank-APGD-100 at $2/255$
and $4/255$ on the attacked queries. Best
value per model block in bold.}\label{tab:budget}
\scriptsize
\setlength{\tabcolsep}{2pt}
\begin{tabular*}{\textwidth}{@{\extracolsep\fill}l ccc ccc ccc ccc ccc@{}}
\toprule
& \multicolumn{3}{c}{MSLS} & \multicolumn{3}{c}{Pitts30k} & \multicolumn{3}{c}{SPED} & \multicolumn{3}{c}{Nordland} & \multicolumn{3}{c}{Mean} \\
\cmidrule(lr){2-4}\cmidrule(lr){5-7}\cmidrule(lr){8-10}\cmidrule(lr){11-13}\cmidrule(lr){14-16}
Method & Clean & 2 & 4 & Clean & 2 & 4 & Clean & 2 & 4 & Clean & 2 & 4 & Clean & 2 & 4 \\
\midrule
\multicolumn{16}{@{}l}{\emph{SuperVLAD}} \\
Plain AT & \textbf{68.9} & 0.0 & 0.0 & \textbf{92.3} & 0.0 & 0.0 & \textbf{90.0} & 0.0 & 0.0 & \textbf{85.7} & 0.0 & 0.0 & \textbf{84.2} & 0.0 & 0.0 \\
ARAFT (Ours) & 59.4 & \textbf{38.0} & \textbf{14.8} & 90.7 & \textbf{76.4} & \textbf{50.1} & 68.9 & \textbf{45.1} & \textbf{22.4} & 68.7 & \textbf{24.7} & \textbf{3.7} & 71.9 & \textbf{46.1} & \textbf{22.8} \\
\midrule
\multicolumn{16}{@{}l}{\emph{BoQ}} \\
Plain AT & \textbf{74.5} & 0.0 & 0.0 & \textbf{93.7} & 0.0 & 0.0 & \textbf{92.6} & 0.0 & 0.0 & \textbf{90.7} & 0.0 & 0.0 & \textbf{87.9} & 0.0 & 0.0 \\
ARAFT (Ours) & 65.1 & \textbf{45.5} & \textbf{21.7} & 91.3 & \textbf{78.8} & \textbf{54.0} & 69.0 & \textbf{47.5} & \textbf{23.6} & 71.9 & \textbf{27.6} & \textbf{4.9} & 74.3 & \textbf{49.8} & \textbf{26.0} \\
\botrule
\end{tabular*}
\end{table*}

\begin{table*}[t]
\caption{Test results without the clean budget, each method at its most robust
checkpoint. Metrics as in Table~\ref{tab:budget}. Best value per model block in
bold.}\label{tab:main}
\scriptsize
\setlength{\tabcolsep}{2pt}
\begin{tabular*}{\textwidth}{@{\extracolsep\fill}l ccc ccc ccc ccc ccc@{}}
\toprule
& \multicolumn{3}{c}{MSLS} & \multicolumn{3}{c}{Pitts30k} & \multicolumn{3}{c}{SPED} & \multicolumn{3}{c}{Nordland} & \multicolumn{3}{c}{Mean} \\
\cmidrule(lr){2-4}\cmidrule(lr){5-7}\cmidrule(lr){8-10}\cmidrule(lr){11-13}\cmidrule(lr){14-16}
Method & Clean & 2 & 4 & Clean & 2 & 4 & Clean & 2 & 4 & Clean & 2 & 4 & Clean & 2 & 4 \\
\midrule
\multicolumn{16}{@{}l}{\emph{SuperVLAD}} \\
Plain AT & 46.2 & 34.9 & \textbf{24.9} & 86.8 & \textbf{77.0} & \textbf{60.8} & 63.1 & \textbf{45.8} & \textbf{29.3} & 50.8 & \textbf{28.0} & \textbf{12.8} & 61.7 & \textbf{46.4} & \textbf{32.0} \\
ARAFT (Ours) & \textbf{59.4} & \textbf{38.0} & 14.8 & \textbf{90.7} & 76.4 & 50.1 & \textbf{68.9} & 45.1 & 22.4 & \textbf{68.7} & 24.7 & 3.7 & \textbf{71.9} & 46.1 & 22.8 \\
\midrule
\multicolumn{16}{@{}l}{\emph{BoQ}} \\
Plain AT & 52.7 & 41.1 & \textbf{26.7} & 88.0 & 78.7 & \textbf{64.0} & 63.3 & 45.1 & \textbf{27.8} & 49.8 & 23.7 & \textbf{7.8} & 63.4 & 47.2 & \textbf{31.6} \\
ARAFT (Ours) & \textbf{65.1} & \textbf{45.5} & 21.7 & \textbf{91.3} & \textbf{78.8} & 54.0 & \textbf{69.0} & \textbf{47.5} & 23.6 & \textbf{71.9} & \textbf{27.6} & 4.9 & \textbf{74.3} & \textbf{49.8} & 26.0 \\
\botrule
\end{tabular*}
\end{table*}

\paragraph{Under the clean budget.} Within the budget, plain adversarial training leaves
the model unchanged and with zero robustness. ARAFT raises attacked R@1 at $2/255$ from 0
to 24.7--76.4 points on SuperVLAD and to 27.6--78.8 points on BoQ, and at $4/255$ to
3.7--50.1 and 4.9--54.0 points (Table~\ref{tab:budget}).

\paragraph{Without the clean budget.} Plain adversarial training
gives up about ten more clean points than ARAFT on average (mean test clean R@1 61.7
vs.\ 71.9 on SuperVLAD and 63.4 vs.\ 74.3 on BoQ) and in exchange is more robust at
$4/255$: by 9.2 mean points on SuperVLAD and 5.6 on BoQ, significant on every dataset
(paired McNemar tests on the same attacked queries, $p\le0.003$). At $2/255$ the two
methods are level on average (46.1 vs.\ 46.4 on SuperVLAD, 49.8 vs.\ 47.2 on BoQ).
Per dataset, ARAFT is significantly better on MSLS for both models ($p=0.018$ and
$p<10^{-3}$) and on Nordland for BoQ ($p=0.004$), plain adversarial training on Nordland
for SuperVLAD ($p=0.016$), and the remaining differences are not significant
($p\ge0.17$).

Rank-PGD-20 over $\eps\in\{1,2,4,8\}/255$ shows the same pattern across budgets. On all eight
model--dataset pairs ARAFT has the higher R@1 at $\eps=0$ and at $1/255$, the two methods
cross near $2/255$, and plain adversarial training is ahead at $4/255$ and $8/255$. At
$8/255$, twice the training budget, ARAFT retains 0.0--6.4 R@1 against 0.4--31.0 for
plain adversarial training. ARAFT's robustness is concentrated at and below the training
budget, where its lower clean cost also lifts the curve.

\paragraph{Transfer to BoQ.} The recipe tuned on SuperVLAD carries over to BoQ unchanged:
its most robust checkpoint again lies inside the budget, the ordering against plain
adversarial training is the same, and ARAFT's mean R@1 is higher than on SuperVLAD at
both budgets (49.8 and 26.0 vs.\ 46.1 and 22.8) for a similar clean drop (13.6 vs.\ 12.3
mean test points).

\subsection{Gradient masking}\label{sec:res_masking}

Four checks on SuperVLAD find no sign of obfuscated gradients in ARAFT.
(i) Rank-APGD with 100 steps and rank-PGD with 20 steps agree within 0.4 R@1 points at
$4/255$ on every dataset. (ii) On MSLS, rank-APGD with 500 steps, or with 100 steps and
three restarts, changes R@1 by 0.00 points relative to the headline attack. (iii) At
$\eps\approx32/255$, R@1 falls to 0.00. (iv) Attacked R@1 never increases with $\eps$.

\subsection{Ablations}\label{sec:res_ablation}

\begin{figure}[t]
\centering
\includegraphics[width=\columnwidth]{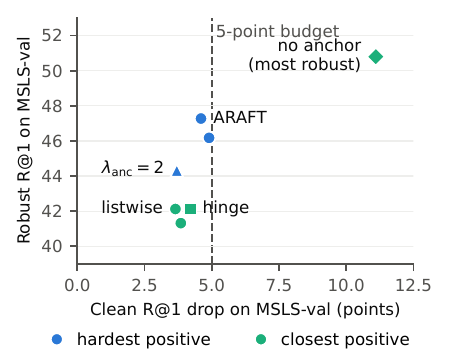}
\caption{Ablations on MSLS-val. Each marker is the budget-5 checkpoint of one run, or for
the run without an anchor, which has none, its most robust checkpoint. Colour gives the
protected positive and the marker the other change: circles are the listwise loss with
$\lambda_{\mathrm{anc}}=1$ (two seeds each), the square the hinge loss, the triangle
$\lambda_{\mathrm{anc}}=2$ and the diamond $\lambda_{\mathrm{anc}}=0$. The robust score
is the mean R@1 under rank-PGD at $2/255$ and $4/255$.}
\label{fig:ablation}
\end{figure}

Table~\ref{tab:ablation} and Fig.~\ref{fig:ablation} vary one component of ARAFT at a
time on SuperVLAD; the figure gives the validation results, the table the test results.

\begin{table*}[t]
\caption{Ablations on SuperVLAD. R@1 under rank-APGD-100 at $4/255$ on the test sets
(clean R@1 in parentheses) of the most robust checkpoint. Validation results are in
Fig.~\ref{fig:ablation}.}\label{tab:ablation}
\footnotesize
\begin{tabular*}{\textwidth}{@{\extracolsep\fill}lccc cc@{}}
\toprule
& & & & \multicolumn{2}{c}{Test, rank-APGD $4/255$} \\
\cmidrule(lr){5-6}
Variant & Positive & Rank loss & $\lambda_{\mathrm{anc}}$ & MSLS & Nordland \\
\midrule
ARAFT & hardest & listwise & 1 & \textbf{14.8} (59.4) & \textbf{3.7} (68.7) \\
Closest positive & closest & listwise & 1 & 10.3 (60.4) & 2.4 (70.4) \\
Closest positive, hinge & closest & hinge & 1 & 10.9 (59.7) & 2.6 (71.4) \\
No anchor & closest & listwise & 0 & 19.8 (50.0) & 8.0 (58.0) \\
Stronger anchor & hardest & listwise & 2 & 11.8 (60.4) & 3.4 (71.0) \\
\botrule
\end{tabular*}
\end{table*}

\paragraph{Hardest positive.} Protecting the hardest instead of the closest positive
raises the validation robust score by 5.2 points, against a seed-to-seed spread of at
most 1.1 points within each variant. The gain holds on the test
sets: at $4/255$ the hardest-positive model is better on MSLS, Pitts30k, SPED and Nordland
(14.8 vs.\ 10.3, 50.1 vs.\ 42.9, 22.4 vs.\ 20.1, 3.7 vs.\ 2.4), significantly so on all
four (paired McNemar, $p\le0.016$), at a clean cost 0.1--2.5 points higher.

\paragraph{Anchor.} Without the anchor, no checkpoint stays within the 5-point budget;
the most robust one reaches a validation robust score of 50.8 at a clean drop of 11.1
points, and loses 18.9 test clean points on MSLS and 27.7 on Nordland. The anchor is
therefore what places the defense inside the budget. Doubling its weight moves the clean
drop from 4.6 to 3.7 validation points but lowers the robust score from 47.3 to 44.3, and
in the short screens (Appendix~\ref{app:search}) weights of 0.1 and 10 were worse than 1.

\paragraph{Rank surrogate.} With the closest positive, the listwise and hinge surrogates
give the same validation robust score (42.1) and test robustness within 0.6 points. We
did not train a hinge variant with the hardest positive, so the surrogate's effect is
established for the closest-positive setting only.

\paragraph{Depth and training attack.} In the short screens, fine-tuning the whole network
was necessary: with 8 or 4 frozen DINOv2 blocks the budget-5 robust score was at most
0.15, against 35.4 with none frozen. Adding two perceptual attacks to the training mix
reduced it from 35.4 to 6.3.

\subsection{Clean cost under distribution shift}\label{sec:res_clean}

The 5-point budget is a rule on the validation set, and the test sets show how far it
transfers (Fig.~\ref{fig:clean_cost}; exact values in Table~\ref{tab:clean} of
Appendix~\ref{app:clean}). On Pitts30k, ARAFT's clean cost is small (1.6 points
on SuperVLAD, 2.5 on BoQ). On MSLS it roughly doubles relative to validation, to 9.4
points on both models, and on SPED and Nordland, whose seasonal and illumination changes are far from
the training data, it reaches 17--24 points. Part of this cost comes from fine-tuning
itself: the clean twin, trained without any attack, already loses 2.8 points on MSLS and
6.5 on Nordland. Relative to the twin, ARAFT's robustness costs 1.6--19.3 clean points.
Plain adversarial training at its most robust checkpoint loses more on every dataset, up
to 34.9 (SuperVLAD) and 41.0 (BoQ) points on Nordland.

\begin{figure}[t]
\centering
\includegraphics[width=\columnwidth]{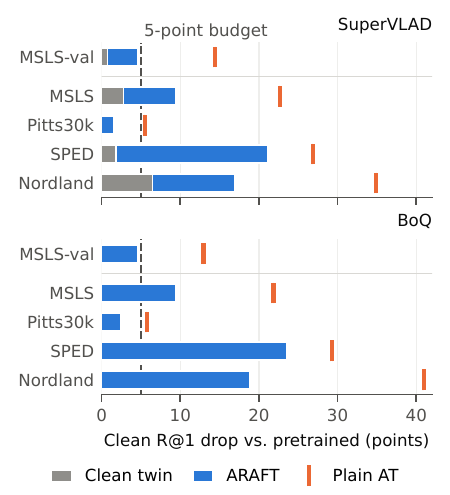}
\caption{Clean R@1 drop relative to the pretrained model on MSLS-val, where the budget
is set, and on the four test sets (all queries). On SuperVLAD the grey part is the drop
of the clean twin, clipped at zero on Pitts30k, where the twin is 0.03 points above the
pretrained model, and the blue part the additional drop of ARAFT; BoQ has no twin.
ARAFT at its budget-5 checkpoint, plain AT (ticks) at its most robust checkpoint.}
\label{fig:clean_cost}
\end{figure}

%% file: sections/06_discussion.tex
\section{Discussion and limitations}\label{sec:discussion}

\paragraph{Why the anchor keeps the defense inside the budget.} The two methods differ
most in the first adversarial epochs: after three epochs, plain adversarial training has
lost 7--10 clean points and ARAFT 1--2 (Fig.~\ref{fig:tradeoff}). A rank
loss only asks the attacked query to stay closer to its positive than to its negatives;
any rearrangement of the descriptor space that achieves this is acceptable to it,
including one that degrades the clean neighbourhoods learned during pretraining. The
anchor removes most of this freedom. By asking the attacked descriptor to coincide with the
pretrained model's clean descriptor, it makes robustness and preservation of the pretrained
geometry the same goal. The ablations agree: without the anchor no checkpoint stays within
the budget, and with it the most robust checkpoint lies inside the budget, or for one
seed 0.4 points outside it, on both architectures. The rank target matters as well:
protecting the hardest positive instead of the closest one improves test robustness on
all four datasets.

\paragraph{Clean cost under distribution shift.} The 5-point budget is enforced on
MSLS-val, and it does not transfer to strongly shifted domains: on SPED and Nordland ARAFT
loses 17--24 clean points. Part of this cost is due to fine-tuning on GSV-Cities at all,
since the clean twin alone loses up to 6.5 points, but most of it comes with robustness.
Because the anchor needs no labels, one possible remedy is to apply it to unlabelled
images of the target domain during fine-tuning. Until such a remedy is shown to work,
practitioners should measure the clean cost on data from their own deployment domain
rather than rely on the validation budget.

\paragraph{Limitations.} Our robustness is empirical. We tested it with several
optimisers and budgets and found no sign of gradient masking, but we did not run
gradient-free or transfer attacks against the ARAFT models, and certified guarantees are
out of scope. Our rank-APGD uses APGD's step-size control without its momentum term;
increasing its steps five-fold or adding restarts did not change the results. The
gradient-masking checks, the clean twin and the second seed are available for SuperVLAD only; BoQ was
trained and evaluated with one seed and the two headline attacks. The threat model covers
digital perturbations of the query; attacks on the database and physical or patch attacks
are not addressed. Finally, robustness at $4/255$ on Nordland is near zero for every
method we trained, so on this benchmark only the $2/255$ results indicate a meaningful
defense.

%% file: sections/07_conclusion.tex
\section{Conclusion}\label{sec:conclusion}

Current foundation-model VPR models lose essentially all top-1 recall under
imperceptible white-box perturbations of the query. We introduced ARAFT, which fine-tunes a released VPR model with a
listwise rank loss on the hardest positive and an anchor to the frozen pretrained model,
and selects checkpoints under an explicit clean-accuracy budget. Both components matter:
the anchor makes robustness and preservation of the pretrained geometry the same goal,
and without it no checkpoint stays within the budget, while protecting the hardest
rather than the closest positive improves test robustness on all four datasets. Within a 5-point
validation budget, ARAFT raises the validation robust R@1 from zero to 47.3 on
SuperVLAD and 55.3 on BoQ, where plain adversarial training has no admissible checkpoint; it carries over to a second
architecture without re-tuning, and its robustness holds against stronger optimisers.
Plain adversarial training exceeds this robustness only at about three times the clean
cost. On the test sets, ARAFT matches plain adversarial training at its most robust
checkpoint at $2/255$ with about ten more clean R@1 points, but falls 6--9 points behind
at $4/255$. Reducing the clean cost under strong
distribution shift, and extending robustness beyond the training budget, are the main
open problems for robust place recognition.

%% file: sections/08_declarations.tex
\section*{Statements and Declarations}
\bmhead{Funding} No funds, grants, or other support were received for this work.
\bmhead{Competing interests} The author has no relevant financial or non-financial
interests to disclose.
\bmhead{Data availability} All datasets analysed in this study are publicly available:
GSV-Cities, MSLS, Pitts30k, SPED and Nordland, from the sources cited in
Section~\ref{sec:data}. The pretrained SuperVLAD and BoQ models are those released by
their authors.
\bmhead{Code availability} The code for training and evaluation is publicly available at
\url{https://github.com/MostafaKhaled2017/vpr-adversarial-robustness}.

%% file: sections/09_appendix.tex
\section{Hyperparameter search}\label{app:search}

All tuning used SuperVLAD and MSLS-val only. Each screen trained for 12 epochs with the
settings of Table~\ref{tab:hparams} and was scored by the robust score of its budget-5
checkpoint (Sect.~\ref{sec:selection}). Stages ran in order, and each carried its winner
forward. The screens used the closest positive as the rank target, which was compared
with the hardest positive only afterwards, in full-length runs.

\begin{table}[h]
\caption{Short screens (12 epochs, closest positive). Clean drop and robust score of the
budget-5 checkpoint on MSLS-val; the winner of each stage in bold.}\label{tab:screens}
\footnotesize
\setlength{\tabcolsep}{3pt}
\begin{tabular}{@{}llcc@{}}
\toprule
Stage & Setting & Drop & Robust \\
\midrule
Attack set & $\ell_\infty$+perc., 8 frozen & $-0.70$ & 0.00 \\
           & $\ell_\infty$, 8 frozen & 0.50 & 0.00 \\
Depth, lr & 8 frozen, $3\cdot10^{-6}$ & 3.00 & 0.03 \\
           & 4 frozen, $10^{-5}$ & 1.05 & 0.15 \\
           & 4 frozen, $3\cdot10^{-6}$ & 0.60 & 0.05 \\
           & \textbf{0 frozen, $10^{-5}$} & 4.25 & \textbf{35.38} \\
           & 0 frozen, $3\cdot10^{-6}$ & 4.90 & 30.90 \\
$\lambda_{\mathrm{anc}}$ & 0 & \multicolumn{2}{c}{none in budget} \\
           & 0.1 & 3.25 & 4.75 \\
           & \textbf{1} & 4.25 & \textbf{35.38} \\
           & 10 & 3.40 & 30.70 \\
Re-check & $\ell_\infty$+perc., 0 frozen & 1.00 & 6.32 \\
\botrule
\end{tabular}
\end{table}

\paragraph{Full-length runs.} With the screened settings fixed, full-length runs compared
the closest and the hardest positive (Table~\ref{tab:ablation}). The hardest positive won
by 5.2 robust points and defines ARAFT. Because it moves the model further from its clean
behaviour, the anchor weight was re-checked with the hardest positive; $\lambda_{\mathrm{anc}}=2$
lowered the robust score (44.3 vs.\ 47.3), and $\lambda_{\mathrm{anc}}=1$ was kept.

\paragraph{Baseline.} Plain adversarial training was screened at learning rates
$10^{-5}$ and $3\cdot10^{-6}$ with no frozen blocks. Neither screen had a checkpoint
within the 5-point budget; at epoch 12 they reached robust scores of 40.75 and 38.90 at
clean R@1 of 65.70 and 67.15. The full-length run uses $10^{-5}$.

\section{Clean recall on the test sets}\label{app:clean}

Table~\ref{tab:clean} gives the values plotted in Fig.~\ref{fig:clean_cost}.

\begin{table}[h]
\caption{Test clean R@1 on all queries, and drops relative to the pretrained model and to
the clean twin (SuperVLAD only).}\label{tab:clean}
\footnotesize
\setlength{\tabcolsep}{2.5pt}
\begin{tabular}{@{}lcccc@{}}
\toprule
& MSLS & Pitts30k & SPED & Nordland \\
\midrule
\multicolumn{5}{@{}l}{\emph{SuperVLAD}} \\
Pretrained & 68.85 & 92.30 & 89.95 & 85.68 \\
Clean twin & 66.05 & 92.33 & 88.14 & 79.20 \\
ARAFT (seed 0) & 59.41 & 90.74 & 68.86 & 68.69 \\
\quad drop vs.\ pretrained & 9.44 & 1.56 & 21.09 & 16.99 \\
\quad drop vs.\ twin & 6.63 & 1.58 & 19.28 & 10.51 \\
Plain AT drop & 22.67 & 5.52 & 26.85 & 34.90 \\
\midrule
\multicolumn{5}{@{}l}{\emph{BoQ}} \\
Pretrained & 74.51 & 93.74 & 92.59 & 90.72 \\
ARAFT & 65.08 & 91.26 & 69.03 & 71.93 \\
\quad drop vs.\ pretrained & 9.43 & 2.48 & 23.56 & 18.79 \\
Plain AT drop & 21.86 & 5.76 & 29.33 & 40.95 \\
\botrule
\end{tabular}
\end{table}